\documentclass[letterpaper]{article} 
\usepackage{aaai2027}  
\usepackage[hyphens]{url}  
\usepackage{graphicx} 
\usepackage{natbib}  
\usepackage{caption} 
\usepackage{algorithm}
\usepackage{algorithmic}
\usepackage{amsmath}
\usepackage{multirow}
\usepackage[table]{xcolor}
\usepackage{tabularx} 

\usepackage{newfloat}
\usepackage{listings}
\DeclareCaptionStyle{ruled}{labelfont=normalfont,labelsep=colon,strut=off} 
\floatstyle{ruled}
\newfloat{listing}{tb}{lst}{}
\floatname{listing}{Listing}

\usepackage{booktabs}

\nocopyright

\title{Zero-Mem: Zero-Token Memory Operations for LLM Agents}
\author{
   Yilin Xiao\textsuperscript{$\spadesuit$}, Zhehan Zhu\textsuperscript{$\clubsuit$}, Yujing Zhang\textsuperscript{$\spadesuit$}, Jin Chen\textsuperscript{$\clubsuit$}, Zijin Hong\textsuperscript{$\spadesuit$}, Luyao Zhuang\textsuperscript{$\spadesuit$},\\ Qinggang Zhang\textsuperscript{$\diamondsuit$}, Shengyuan Chen\textsuperscript{$\spadesuit$}, Xiaocao Ouyang\textsuperscript{$\clubsuit$}, Lingfei Ren\textsuperscript{$\clubsuit$}, Xiao Huang\textsuperscript{$\spadesuit$}   
}
\affiliations{
    \textsuperscript{$\spadesuit$}The Hong Kong Polytechnic University, Hong Kong SAR\\
    \textsuperscript{$\clubsuit$}School of Computing and Artificial Intelligence,\\ Southwestern University of Finance and Economics\\
    \textsuperscript{$\diamondsuit$}School of Artificial Intelligence, Jilin University\\
     \texttt{yilin.xiao@connect.polyu.hk} \hspace{1.0em}
    \texttt{42436006@smail.swufe.edu.cn} \\
    \texttt{renlf@swufe.edu.cn} \hspace{1.0em} \texttt{xiao.huang@polyu.edu.hk} \\

}

\begin{document}

\maketitle

\begin{abstract}
LLM agents need memory to act consistently over long interactions, yet many systems use additional LLM calls to operate that memory. Generating intermediate records and mediating their retrieval adds recurring token and time costs, while omitted or merged details can obscure the original evidence. We ask whether structured memory access requires generation at all. Zero-Mem introduces \emph{zero-token memory operations}: no step outside final question answering invokes an LLM or consumes LLM input or output tokens; encoder computation is accounted for separately. Zero-Mem preserves original interaction traces as its source of record. It organizes the traces in two complementary ways. An entity--context graph exposes connections across interactions, while a temporal hierarchy preserves conversational locality and session state. For each query, Zero-Mem weighs the two views, retrieves from both, and follows their structure to recover supporting relations or surrounding context. Deterministic calibration first discards conflicting evidence and then keeps the reader's answer grounded in the retrieved traces. Only the final-QA reader invokes an LLM. Across long-memory and long-context question-answering benchmarks, Zero-Mem achieves competitive performance while eliminating LLM calls and LLM-token consumption from memory operations. With the same final-QA reader and context budget, it reduces memory-operation time cost by 57.6\% relative to the fastest compared baseline. Ablations support the contribution of the two views and their query-dependent coordination. Overall, the results show that structured agent memory need not generate an intermediate representation of the past. The code and implementation details are available at \textcolor{blue}{\url{https://github.com/Zero-Mem/Zero-mem}}.
\end{abstract}


\section{Introduction}
Large language model (LLM) agents increasingly operate over extended interactions, accumulating utterances, actions, tool observations, and task outcomes~\cite{agent_survey_1,agent_survey_2,agent_survey_3}. Their reliability therefore depends not only on reasoning over the current input, but also on recovering the right evidence from a growing interaction history. A memory system must preserve information across sessions while preventing irrelevant or outdated traces from dominating the current decision. The central challenge is thus no longer merely how to store more context, but how to recover evidence associated with the correct entity, session, and temporal state when it becomes relevant~\cite{memory_survey_1,memory_survey_2,memory_survey_3,memory_survey_4}.

\begin{figure}[t]
    \centering
    \includegraphics[width=\linewidth]{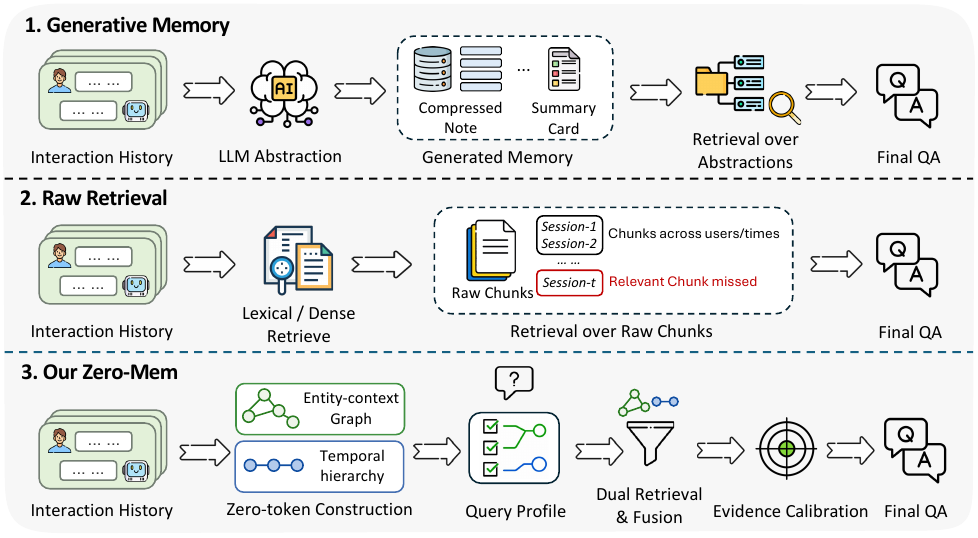}
    \caption{Comparison of different agent-memory operating regimes. Generative memory relies on LLM-generated abstractions, while raw retrieval searches unstructured traces and may miss distributed evidence. Zero-Mem builds relational and temporal memory structures and performs all memory operations with zero LLM calls or tokens; only final QA invokes an LLM.}
    \label{fig:intro}
\end{figure}

Across agent-memory and agentic structured-retrieval systems, language models have been used to summarize or reflect on experience, construct hierarchical abstractions and graph indexes, and generate or evolve linked memory records~\cite{embedding_4,embedding_1,embedding_3,graph_2,graph_3,graph_4}. These transformations can make large histories easier to access, but they also turn memory management into a recurring generative workload. When generated abstractions mediate later retrieval, omitted details, merged subjects, or blurred temporal updates may weaken traceability to the original interaction. The opposite strategy is to retain the complete history and retrieve directly from raw traces~\cite{GAM, raw_1}. Although this preserves source evidence, flat lexical or dense retrieval can confuse semantically similar traces from different users, sessions, or temporal states, and may fail when supporting evidence is distributed across multiple interactions. Effective memory therefore requires faithful preservation and structured, query-conditioned evidence selection.

Recent systems reduce this dependence rather than eliminate it. SimpleMem~\cite{simplemem} improves token efficiency through semantic structured compression, online semantic synthesis, and intent-aware retrieval planning, while LightMem~\cite{LightMem} shifts several memory operations from large LLMs to specialized small language models and separates online retrieval from offline consolidation. These approaches reduce generative overhead, but do not target a memory pipeline in which final question answering is the only LLM-dependent stage. We therefore ask: \textit{Can an agent memory system eliminate LLM calls from every operation outside final question answering, while retaining structured access beyond flat similarity retrieval?} We refer to this operating regime as \textbf{zero-token memory operations}: memory construction, organization, routing, retrieval, evidence closure, and both pre-reader evidence calibration and post-reader answer calibration invoke no LLM and consume no LLM input or output tokens. Encoder computation and final-QA inference are accounted for separately.

We propose \textbf{Zero-Mem}, which reformulates memory operation as structured evidence selection over provenance-bearing interaction traces. Rather than replacing histories with generated abstractions, Zero-Mem retains the original traces as the source of record and derives two complementary, non-generative views over them. An entity--context graph captures observed co-occurrence and trace adjacency for relational access, while a temporally ordered hierarchy preserves conversational locality and session-level state. Both views resolve to the same provenance-bearing source units. At query time, a lightweight profile coordinates the two views according to the structural requirements of the query. Their rankings are fused, and evidence closure supplements the main candidates with relational connections and surrounding trace context. Deterministic evidence calibration then produces a compact evidence set $R(q)$ for final QA. The reader is the only LLM-dependent stage; afterward, deterministic answer calibration applies evidence-support, type, and format checks without invoking another model. Thus, no generated memory intervenes between the original trace and the evidence exposed to the reader.

Across the long-context and long-memory QA benchmarks, Zero-Mem achieves competitive performance while reducing memory-operation LLM calls and tokens to zero. With an identical final-QA reader and equivalent context budget, Zero-Mem achieves a 57.6\% reduction in memory-operation latency compared to the most time-efficient baseline, and ablation studies further verify the effectiveness of each core module. Our contributions are threefold:

\begin{itemize}
    \item We define \textbf{zero-token agent memory}, an operating regime in which every operation outside final QA uses zero LLM calls and zero LLM input or output tokens, separating memory-operation cost from final-reader inference.

    \item We introduce \textbf{Zero-Mem}, a provenance-preserving framework that coordinates relational and temporally ordered views to perform structured evidence selection directly over original interaction traces.

    \item We evaluate Zero-Mem on multiple long-memory benchmarks, demonstrating its competitive performance under zero memory-operation LLM cost and analyzing the contributions of its complementary core modules.
\end{itemize}

\section{Related Work}
Agent-memory systems organize, update, and retrieve growing interaction histories~\cite{related_1,related_2,related_3,related_4,related_5}. Zep~\cite{zep} builds a temporally aware knowledge-graph memory layer with episodic, semantic-entity, and community subgraphs and a dual-time model tracking event and ingestion times. Mem0~\cite{Mem0} incrementally extracts and updates memories through LLM tool calls for add, update, delete, and no-op operations; Mem0g models entity relations with a directed labeled graph. A-Mem~\cite{A-Mem} follows the Zettelkasten note-taking method, constructing structured memory notes with keywords, tags, and contextual descriptions while dynamically linking related memories. MemoryOS~\cite{MemoryOS} uses an operating-system-inspired architecture with short-term, mid-term, and long-term storage, paging, and popularity-based updates. GAM~\cite{GAM} combines lightweight offline memory with online deep research under a just-in-time memory paradigm, constructing task-specific contexts at higher query-time cost. CompassMem~\cite{CompassMem} organizes experiences into event-centric memory graphs with explicit relations for complex questions. LightMem~\cite{LightMem} decouples memory updates from online inference, applying pre-compression and topic segmentation to reduce latency and token cost. SimpleMem~\cite{simplemem} combines semantic structured compression, online semantic synthesis, and intent-aware retrieval planning to reduce token consumption. Together, these systems improve memory efficiency, while many retain generative processing within the memory lifecycle.

\section{Preliminaries}
An LLM agent accumulates a history of past interactions
$\mathcal{H}=(s_1,\ldots,s_T)$, where each trace unit $s_i$ may contain user messages, assistant responses or actions, tool observations, timestamps, speakers, and session metadata. Given a current query $q$, an agent memory system retrieves relevant information from the history to construct an evidence set
\begin{equation}
R(q)=\operatorname{Memory}(q,\mathcal{H}).
\end{equation}
A reader LLM then uses the retrieved evidence to produce the answer:
\begin{equation}
a=\operatorname{Reader}(q,R(q)).
\end{equation}
In this work, Zero-Mem instantiates the memory function through non-generative memory construction, organization, retrieval, routing, and calibration.


\begin{figure*}[t]
    \centering
    \includegraphics[width=\linewidth]{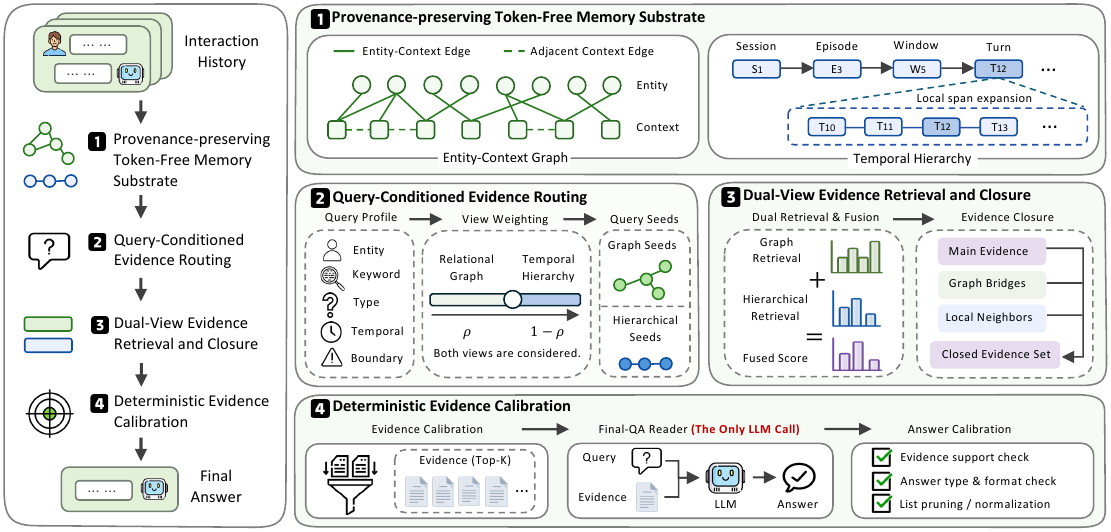}
    \caption{Overview of Zero-Mem. Original interaction traces are organized into a provenance-preserving entity--context graph and temporal hierarchy without generative abstraction. Query-conditioned routing weights the two views, whose retrieved evidence is fused and completed with relational bridges and local neighbors. Deterministic calibration filters and ranks the evidence and checks the reader output. All memory operations are token-free; the final-QA reader is the only LLM call.}
    \label{fig:main_figure}
\end{figure*}

\section{Method}

\subsection{Overview of Zero-Mem}

Zero-Mem implements the memory function through token-free evidence selection. It retains original interaction traces as the authoritative memory source and builds non-generative retrieval structures over them. Zero-Mem consists of four components: a \textbf{Provenance-preserving Token-Free Memory Substrate}, \textbf{Query-Conditioned Evidence Routing}, \textbf{Dual-View Evidence Retrieval and Closure}, and \textbf{Deterministic Evidence Calibration}. The graph view recovers relational evidence, while the hierarchical view preserves local, temporal, and session context. Routing controls their relative priority, closure supplements the retrieved candidates with structurally related evidence, and calibration removes inconsistent or unsupported content. All memory operations are token-free, and only the final reader produces the answer.

\subsection{Provenance-preserving Token-Free Memory Substrate}

Zero-Mem does not replace raw histories with generated abstractions. Each derived unit retains its original text together with source identifier, session time, boundary identifier, and other available metadata. Consequently, retrieved evidence remains traceable to observed interactions rather than model-generated memory statements.

\paragraph{Relational trace graph.}
Zero-Mem applies the non-generative Named Entity Recognition (NER) model (e.g., spaCy) to each context unit and constructs an observed entity--context graph from the detected entities:
\begin{equation}
G=(V_d\cup V_e,E_{de}\cup E_{dd}),
\end{equation}
where $V_d$ and $V_e$ denote context and entity nodes, respectively. $E_{de}$ contains entity--context co-occurrence edges, and $E_{dd}$ contains adjacency edges between neighboring context units. An entity--context edge is added when entity $e$ is detected in context unit $d_i$, with weight:
\begin{equation}
    w(d_i,e)=
    \frac{c(e,d_i)}
    {\sum_{e'\in\mathcal{E}(d_i)}c(e',d_i)}.
\end{equation}

where $c(e,d_i)$ is the occurrence frequency of $e$ in $d_i$. $\mathcal{E}(d_i)$ denotes the set of entities detected in $d_i$. Adjacent context units are also connected to preserve local continuity. The graph records observed co-occurrence and trace adjacency rather than generating semantic triples or inferred relations.

\paragraph{Hierarchical trace units.}
Graph structure alone does not preserve the local order and temporal state of an interaction. Zero-Mem organizes traces at multiple granularities:

\begin{equation}
    \mathcal{T}(\mathcal{H})
    =
    U_{\mathrm{turn}}
    \cup U_{\mathrm{window}}
    \cup U_{\mathrm{episode}}
    \cup U_{\mathrm{local}}.
\end{equation}

Turns preserve atomic utterances, windows retain short-range context, and episodes group adjacent windows into coherent event regions according to semantic continuity and available temporal or session boundaries. Local spans preserve the immediate neighborhood of a candidate turn and are used when the selected evidence requires surrounding context. All units inherit provenance from their underlying raw traces.

\paragraph{Lexical and dense access signals.}
Zero-Mem additionally indexes trace units with lexical statistics (BM25) and dense embeddings (BGE-M3). Lexical signals identify exact names, dates, numbers, titles, and phrases, while dense signals provide semantic anchors when surface overlap is weak. These representations support indexing, seeding, and scoring only; they do not generate or rewrite memory content.

\subsection{Query-Conditioned Evidence Routing}

For each query, Zero-Mem constructs a lightweight profile
\begin{equation}
\begin{aligned}
\phi(q)=\{&
\mathrm{subject},\mathrm{keywords},
\mathrm{answer\mbox{-}type},\\
&\mathrm{temporal\mbox{-}cues},
\mathrm{boundary}\}.
\end{aligned}
\end{equation}

The subject and keywords provide content anchors, while the answer type and temporal cues characterize the structural requirements of the requested evidence. When available, the boundary specifies the admissible interaction scope. These signals are obtained from the query and available metadata without using gold answers, and are shared by routing and subsequent evidence selection. The profile determines which evidence view receives priority:

\begin{equation}
    \operatorname{Route}(q)\in
    \{\mathrm{relational},\mathrm{local}\}.
\end{equation}

The relational route denotes graph priority, whereas the local route denotes hierarchy priority. The routing decision is based on deterministic query-structure signals, including question form, temporal or aggregation requirements, and the availability of subject anchors. Both views are executed in the full model; routing primarily controls their relative weights during fusion. Let $\rho$ denote the globally shared primary-view weight. Relational queries assign weights $\rho$ and $1-\rho$ to the graph and hierarchical views, respectively, while local queries reverse these weights.

\subsection{Dual-View Evidence Retrieval}
\paragraph{Graph evidence propagation.}
The graph view first aligns each entity $\hat e$ extracted from the query with its most similar observed graph entity $e$. Its initial activation is:

\begin{equation}
\eta_0(e\mid q)
=
\cos(\mathbf{e},\hat{\mathbf{e}}),
\qquad
e=
\arg\max_{e'\in V_e}
\cos(\mathbf{e}',\hat{\mathbf{e}}),
\end{equation} where $\mathbf{e}$ and $\hat{\mathbf{e}}$ are their dense representations. Dense context matches provide context priors when aligned entities are available and a direct fallback ranking when none is detected, while lexical and phrase signals refine the resulting context ranking. Zero-Mem then expands activation from these matched graph entities through relevant co-occurrence sentences. Let $Z(e)$ denote the set of sentences containing entity $e$. The propagated activation of entity $e'$ is
\begin{equation}
    \eta_{t+1}(e')
    =
    \sum_{e\in\mathcal{E}_t}
    \eta_t(e)
    \sum_{z\in Z(e)\cap Z(e')}
    \operatorname{sim}(q,z),
\end{equation} where $t$ is the propagation step, $\mathcal{E}_t$ is the set of active graph entities at step $t$, with $\mathcal{E}_0$ consisting of the matched entities, and $\operatorname{sim}(q,z)$ denotes the dense similarity between query $q$ and sentence $z$. An entity therefore receives a high score when it co-occurs with an already activated graph entity in sentences relevant to the query. The propagated entity activations and dense context priors are combined into a query-specific reset vector $\mathbf{r}_q$. Personalized PageRank then distributes this evidence over the relational graph: 
\begin{equation}
\boldsymbol{\pi}_q
=
(1-\gamma)\mathbf{r}_q
+
\gamma P^\top\boldsymbol{\pi}_q,
\end{equation}
where $\boldsymbol{\pi}_q$ is the query-conditioned stationary node-score vector, $\mathbf{r}_q$ is a normalized reset distribution combining propagated entity activations and dense context priors, $P$ is the row-normalized graph transition matrix, and $\gamma\in(0,1)$ is the damping factor. PageRank values on context nodes form the graph-view ranking. Exact lexical and phrase matches are finally used to refine this ranking for names, dates, values, titles, and quoted expressions.

\paragraph{Hierarchical evidence retrieval.}
The hierarchical view retrieves evidence through coarse-to-fine search. Each unit is evaluated by jointly considering its semantic relevance to the query and its structural compatibility with the query profile. The compatibility signals include subject consistency, temporal validity, boundary consistency, expected answer type, and lexical or phrase support. These signals are used to refine the semantic ranking rather than being treated as independently generated evidence. Retrieval proceeds from episodes to windows and then to individual turns:
\begin{equation}
U_{\mathrm{episode}}
\rightarrow
U_{\mathrm{window}}
\rightarrow
U_{\mathrm{turn}}
\rightarrow
U_{\mathrm{local}}.
\end{equation}
Episodes identify relevant event regions, windows narrow the search to local contexts, and turns expose the original evidence. When a selected turn depends on nearby information, its local span is added to preserve the immediate narrative or conversational state. Unlike graph propagation, this view explicitly maintains ordering, temporal locality, and session-level context.

\begin{table*}[t]
\centering
\resizebox{\linewidth}{!}{
\begin{tabular}{clcccccccccc} 
\toprule
\multirow{3}{*}{Model} & \multirow{3}{*}{Method} & \multicolumn{10}{c}{LoCoMo} \\
\cmidrule(lr){3-12}
& & \multicolumn{2}{c}{Single Hop} & \multicolumn{2}{c}{Multi Hop} & \multicolumn{2}{c}{Temporal} & \multicolumn{2}{c}{Open Domain} & \multicolumn{2}{c}{Average} \\
\cmidrule(lr){3-4} \cmidrule(lr){5-6} \cmidrule(lr){7-8} \cmidrule(lr){9-10} \cmidrule(lr){11-12}
& & F1 & BLEU-1 & F1 & BLEU-1 & F1 & BLEU-1 & F1 & BLEU-1 & F1 & BLEU-1 \\
\midrule
\multirow{11}{*}{\rotatebox{90}{\makebox[1em]{GPT-4o-mini}}}
& LONG-LLM   & 46.68 & 37.54 & 29.23 & 22.76 & 25.97 & 19.42 & 16.87 & 13.70 & 37.31 & 29.57 \\
& RAG        & 52.45 & 47.94 & 27.50 & 20.13 & 46.07 & 40.35 & 23.23 & 17.94 & 44.73 & 39.40 \\ \cmidrule{2-12}
& HippoRAG        &  54.84         & 48.84 & 33.59& 25.46 &48.17 & 39.32 & 28.59 &  23.89 &  47.92 & 41.02 \\
& A-Mem      & 44.65 & 37.06 & 27.02 & 20.09 & 45.85 & 36.67 & 12.14 & 12.00 & 39.65 & 32.31 \\
& Mem0       & 47.65 & 38.72 & 38.72 &27.13 & 48.93 & 40.51 & 28.64 & 21.58 & 45.10 & 35.90 \\
& MemoryOS   & 48.62 & 42.99 & 35.27 & 25.22 & 41.15 & 30.76 & 20.02 & 16.52 & 42.84 & 35.54 \\
& LightMem   & 41.79 & 37.83 & 29.78 & 24.80 & 43.71 & 39.72 & 16.89 & 13.92 & 38.44 & 34.36 \\
& SimpleMem   & 53.48 & 47.59 & 36.93 & 28.80 & 51.30 & 45.29 & 21.78 & 15.67 & 48.02 & 41.68 \\
& CompassMem   &  57.36 & 49.79 & 38.84 & 27.98 &  57.96&  50.51 &  26.61  & 20.01 &  52.18 & 44.09 \\
& GAM        & \underline{57.75} & \underline{52.10} & \textbf{42.29} & \textbf{34.44} & \underline{59.45} & \underline{53.11} & \underline{33.30}& \underline{26.97} & \underline{53.75} & \underline{47.51} \\ \cmidrule(lr){2-12}
& Zero-Mem & \textbf{66.65} & \textbf{60.53} & \underline{41.61} & \underline{32.92} & \textbf{61.97} & \textbf{57.45} & \textbf{35.52} & \textbf{30.47} & \textbf{59.15} & \textbf{52.96} \\
\midrule
\multirow{11}{*}{\rotatebox{90}{\makebox[1em]{Qwen2.5-14B}}}
& LONG-LLM   & 46.05 & 39.56 & 32.08 & 24.46 & 30.51 & 24.45 & 14.89 & 11.41 & 38.31 & 31.90 \\
& RAG        & 47.87 & 42.79 & 26.38 & 19.54 & 30.78 & 25.97 & 14.16 & 10.52 & 38.27 & 33.01 \\ \cmidrule{2-12}
& HippoRAG        &  42.45 & 37.14 & 27.57& 20.62 & 30.66 & 26.33 & 19.74  &  15.81 &   35.85  & 30.53 \\
& A-Mem      & 33.75 & 30.04 & 22.09 & 15.28 & 27.19 & 22.05 & 13.49 & 10.74 & 28.98 & 24.47 \\
& Mem0       & 42.58 & 35.15 & 31.73 & 24.82 & 28.96 & 26.24 & 15.03 & 11.28 & 36.04 & 29.91 \\
& MemoryOS   & 46.33 & 41.62 & 38.19 & 29.26 & 32.24 & 27.86 & 20.27 & 15.94 & 40.28 & 34.89 \\
& LightMem   & 34.92 & 31.22 & 25.45 & 19.61 & 32.03 & 27.70 & 15.81 & 11.81 & 31.39 & 27.15 \\
& SimpleMem   & 51.11 & 45.47 & 34.04 & 25.65 & 48.46 & 36.07 & 23.41 & 21.45 & 45.71 & 38.39 \\
& CompassMem   &  \underline{61.02}  & \underline{55.93} & 42.32 & 32.66 & 47.18 &  39.69 &  25.88   & 22.01 &  52.52 &  46.17 \\
& GAM        & 58.93 & 53.74 & \underline{42.96} & \underline{34.48} & \underline{51.52} & \underline{44.43} & \underline{30.63} & \underline{26.04} & \underline{52.70} & \underline{46.55} \\ \cmidrule(lr){2-12}
& Zero-Mem  &\textbf{64.09}  & \textbf{58.19} & \textbf{44.06} & \textbf{35.13} & \textbf{58.34} & \textbf{53.61}& \textbf{37.57}&\textbf{32.46}& \textbf{57.57}& \textbf{51.41} \\
\bottomrule
\end{tabular}
    }
\caption{Performance comparison on LoCoMo. Results are reported across four question types under two evaluation metrics, F1 and BLEU-1, using GPT-4o-mini and Qwen2.5-14B as base LLMs. The best results are shown in \textbf{bold}, and the second-best results are \underline{underlined}.}
\label{tab:locomo_results_weighted}
\end{table*}

\subsection{Dual-View Evidence Closure}

Zero-Mem first aligns the graph and hierarchical rankings through
query-wise score normalization. For each view $v\in\{g,h\}$,
\begin{equation}
\widehat{S}_v(d)=
\begin{cases}
0,
& d\text{ is absent from view }v,\\[2pt]
\dfrac{S_v(d)-S_v^{\min}}
{S_v^{\max}-S_v^{\min}},
& S_v^{\max}>S_v^{\min},\\[8pt]
1,
& S_v^{\max}=S_v^{\min},
\end{cases}
\end{equation}
where $S_v^{\min}$ and $S_v^{\max}$ are computed over the candidates
returned by view $v$. The normalized rankings are fused using the dual-view routing coefficient $\rho$:
\begin{equation}
S_{\mathrm{fuse}}(d)
=
\rho\,\widehat{S}_{\mathrm{primary}}(d)
+
(1-\rho)\,\widehat{S}_{\mathrm{secondary}}(d).
\end{equation}
The graph view is primary for relational queries, whereas the
hierarchical view is primary for local queries. Let $M(q)$ denote the main evidence retained after fusion. Zero-Mem augments it with bounded, query-conditioned support from the two views:
\begin{equation}
C(q)=
\operatorname{Dedup}
\left(
M(q)
\cup
\mathcal{N}_g(M(q))
\cup
\mathcal{N}_h(M(q))
\right).
\end{equation}
Here, $\mathcal{N}_g$ supplies additional graph-ranked contexts with relational or bridging support, while $\mathcal{N}_h$ restores neighboring turns or local spans; either support set may be empty when no addition is required. Duplicates are merged using shared unit identifiers or source provenance when available, yielding a compact evidence set with relational and local support.

\subsection{Deterministic Evidence Calibration}

Zero-Mem applies deterministic calibration at both the evidence and answer levels. After evidence closure, it removes candidates that violate provenance or query-boundary constraints and ranks the remaining evidence by subject, temporal, and answer-type compatibility:
\begin{equation}
R(q)=
\operatorname{Rank}_{\phi(q)}
\left(
\operatorname{Filter}
\left(
C(q),\phi(q)
\right)
\right).
\end{equation}
Here, $\operatorname{Filter}$ enforces the hard constraints, whereas $\operatorname{Rank}_{\phi(q)}$ orders the admissible evidence without altering its content. The reader produces an initial answer $a_0$ from $R(q)$. For answer forms admitting deterministic checks, Zero-Mem extracts evidence-local candidates and calibrates the output:
\begin{equation}
\begin{aligned}
A(q)
&=
\operatorname{Extract}
\left(
R(q),\phi_{\mathrm{type}}(q)
\right),\\
a
&=
\operatorname{Calibrate}
\left(
a_0,q,A(q),R(q),\phi(q)
\right).
\end{aligned}
\end{equation}
Calibration preserves $a_0$ when it is supported and well-formed; otherwise, it applies evidence-preserving normalization, extractive shortening, or item-wise list pruning. A scalar answer is replaced only by a unique type-compatible candidate in $A(q)$; if no deterministic correction is available, $a_0$ is retained.

\begin{table*}[t]
\centering
\begin{tabular*}{\textwidth}{
    @{\extracolsep{\fill}}
    lcccccc
    @{}
}
\toprule
\multirow{2}{*}{Method}
& \multicolumn{2}{c}{Answer Quality}
& \multicolumn{4}{c}{Memory-Operation Overhead} \\
\cmidrule(lr){2-3}
\cmidrule(lr){4-7}
& F1 Score
& BLEU-1
& Tokens
& Tokens / Query
& Time (s)
& Time / Query (s) \\
\midrule

SimpleMem
& 48.02
& 41.68
& 14,096,246
& 9,153.41
& 8,365.38
& 5.43 \\

LightMem
& 38.44
& 34.36
& \underline{877,086}
& \underline{569.54}
& \underline{788.76}
& \underline{0.51} \\

GAM
& \underline{53.75}
& \underline{47.51}
& 28,570,674
& 18,552.39
& 9,237.25
& 6.00 \\

\midrule
\textbf{Zero-Mem (Ours)}
& \textbf{59.15}
& \textbf{52.96}
& \textbf{0}
& \textbf{0}
& \textbf{334.77}
& \textbf{0.22} \\

\midrule
Relative Gain/Reduction
& \textbf{10.0\%$\uparrow$}
& \textbf{11.5\%$\uparrow$}
& \textbf{100.0\%$\downarrow$}
& \textbf{100.0\%$\downarrow$}
& \textbf{57.6\%$\downarrow$}
& \textbf{57.6\%$\downarrow$} \\
\bottomrule
\end{tabular*}

\caption{Efficiency comparison under a unified experimental configuration.
All methods use four concurrent threads, GPT-4o-mini as the backbone LLM,
and identical test hardware. Relative Gain/Reduction is computed against
the underlined result.}
\label{tab:quality_efficiency}
\end{table*}







\section{Experiment}
\subsection{Experimental Setup}
\subsubsection{Datasets.}
We evaluate Zero-Mem on two complementary benchmarks. 1) \textbf{LoCoMo}~\cite{locomo} is a widely adopted benchmark for assessing long-term memory in conversational agents over extended, multi-session interactions. Following prior work~\cite{GAM}, we evaluate its single-hop, multi-hop, temporal-reasoning, and open-domain tasks. 2) \textbf{HotpotQA}~\cite{HotpotQA} is a Wikipedia-based benchmark for multi-hop question answering. Following MemAgent~\cite{MemAgent}, we adopt the curated memory-evaluation variant, which combines gold supporting documents with distractor passages. Varying the number of distractors produces three context-length settings of 56K, 224K, and 448K tokens.

\subsubsection{Baselines.}
We organize the comparison methods into two groups. 1) \textbf{Memory-free baselines} comprise LONG-LLM and RAG. LONG-LLM partitions the interaction history into multiple text blocks using a sliding window, processes each block independently, and returns the candidate answer with the highest confidence. RAG divides the history into 2,048-token chunks and retrieves the top five chunks by semantic similarity as supporting context for answer generation. 2) \textbf{Memory-based baselines} comprise A-Mem~\cite{A-Mem}, Mem0~\cite{Mem0}, MemoryOS~\cite{MemoryOS}, LightMem~\cite{LightMem}, SimpleMem~\cite{simplemem}, CompassMem~\cite{CompassMem}, and GAM~\cite{GAM}. These methods maintain specialized memory structures over historical information and access them during inference to support memory-grounded tasks. Additional baseline descriptions are provided in the Appendix.

\subsubsection{Implementation Details.}
We use GPT-4o-mini and Qwen2.5-14B-Instruct as the backbone LLMs for Zero-Mem and all baselines, representing closed-source and open-source settings, respectively. Within each setting, all methods use an identical final-QA reader and equivalent context budget, so the comparison isolates differences in their memory pipelines. Damping factor $\gamma$ and dual-view routing coefficient $\rho$ are both set to 0.6. All experiments are executed in a common hardware environment equipped with NVIDIA RTX 4090 GPUs. For controlled comparison, we cap the number of retrieved items at five for every method. We follow the evaluation metrics and protocols established in prior work~\cite{GAM,simplemem}.

\begin{table}[h]
  \centering
   \small
\setlength{\tabcolsep}{1.7mm}
  \begin{tabular}{lcccccc}
    \toprule
    \multirow{2}{*}{\textbf{Method}} & \multicolumn{3}{c}{\textbf{GPT-4o-mini}} & \multicolumn{3}{c}{\textbf{Qwen2.5-14B}} \\
    \cmidrule(lr){2-4} \cmidrule(lr){5-7}
    & \textbf{56K} & \textbf{224K} & \textbf{448K} & \textbf{56K} & \textbf{224K} & \textbf{448K} \\
    \midrule
    LONG-LLM & 56.56 & 54.29 & 53.92             & 49.75             & 46.82 & 43.17 \\
    RAG      & 52.71             & 51.84             & 54.01 & 51.81 & 46.72             & 48.36 \\
    \midrule
    A-Mem    & 33.90             & 30.22             & 31.37             & 27.04             & 25.65             & 22.92 \\
    Mem0     & 32.58             & 31.74             & 27.41             & 30.12             & 32.44             & 26.55 \\
    MemoryOS & 26.47             & 23.10             & 24.16             & 24.58             & 30.25             & 23.13 \\
    LightMem & 40.93             & 35.28             & 30.02             & 37.30             & 27.72             & 28.25 \\
    GAM      & \underline{63.22}    & \underline{64.56}    & \underline{59.81}    & \underline{64.07}  & \underline{55.99}   & \underline{57.87} \\ \midrule 
   Zero-Mem     & \textbf{72.07}    & \textbf{66.43}    & \textbf{65.04}    & \textbf{68.58}    & \textbf{65.47}    & \textbf{61.02} \\
    
    \bottomrule
  \end{tabular}
  \caption{Performance comparison (F1 score) on HotpotQA across different base LLMs and context length settings.}
  \label{tab:hotpotqa_horizontal_wide}
\end{table}

\subsection{Main Results}
\subsubsection{LoCoMo.}


Table~\ref{tab:locomo_results_weighted} reports the results on LoCoMo, which evaluates memory over multi-session conversations and emphasizes the recovery of entity-specific, temporal, and cross-session information. Zero-Mem achieves the best average F1 and BLEU-1 under both LLM readers. Relative to GAM, the strongest overall baseline, it improves average F1 and BLEU-1 by 5.40 and 5.45 points with GPT-4o-mini, and by 4.87 and 4.86 points with Qwen2.5-14B, respectively. With GPT-4o-mini, Zero-Mem leads on single-hop, temporal, and open-domain questions while remaining competitive with GAM on multi-hop questions. With Qwen2.5-14B, it ranks first across every question type and metric. The sizable margins over LONG-LLM and RAG, particularly on temporal and open-domain questions, indicate that long-context access or flat similarity retrieval alone is insufficient for state- and boundary-sensitive recall. This consistency across LLM readers and memory requirements demonstrates that Zero-Mem can recover relevant conversational evidence while preserving its relational and temporal context, despite requiring no LLM calls or tokens for memory operations.

\subsubsection{HotpotQA.}
Table~\ref{tab:hotpotqa_horizontal_wide} reports the HotpotQA results as the context length increases from 56K to 448K tokens. By progressively adding distracting passages, this benchmark tests whether a method can locate and connect distributed supporting evidence under increasingly long contexts. Zero-Mem achieves the highest F1 across all readers and context lengths, including the challenging 448K-token setting, with an average improvement of 5.52 points over the strongest baseline. Together, the results on LoCoMo and HotpotQA show that Zero-Mem is effective for both long-term conversational memory and long-context multi-hop retrieval, demonstrating the generality of its structured evidence-selection framework under zero-token memory operations.

\subsection{Efficiency Comparison}
Table~\ref{tab:quality_efficiency} evaluates whether reducing memory-operation overhead comes at the expense of answer quality. We compare Zero-Mem with GAM, the strongest-performing baseline in the main experiments, as well as SimpleMem and LightMem, two representative efficiency-oriented memory systems. All methods are evaluated using GPT-4o-mini under the same concurrency setting and hardware environment. We report both answer quality and the total and per-query overhead incurred by memory operations outside the shared final-QA stage. Zero-Mem achieves the highest F1 and BLEU-1 scores, improving them by 10.0\% and 11.5\%, respectively, over GAM, the second-best method on both metrics. Thus, eliminating LLM-based memory operations does not compromise answer quality. In terms of overhead, Zero-Mem invokes no LLM during memory processing and consequently consumes zero LLM input or output tokens, whereas even LightMem, the most token-efficient baseline, consumes more than 0.87 million tokens. Zero-token operation does not imply zero computation, since encoder inference, memory organization, retrieval, and deterministic calibration still incur processing costs. Nevertheless, Zero-Mem requires only 334.77 seconds in total and 0.22 seconds per query, reducing memory-operation latency by 57.6\% relative to LightMem, the fastest baseline. This result indicates that the removal of generative memory calls does not shift the cost to a slower non-generative pipeline. Under the unified setting, Zero-Mem outperforms every compared baseline in answer quality while also achieving the lowest memory-operation token and overhead. These demonstrate that its efficiency gains do not come at the expense of answer quality.

\begin{figure}[t]
    \centering
    \includegraphics[width=\linewidth]{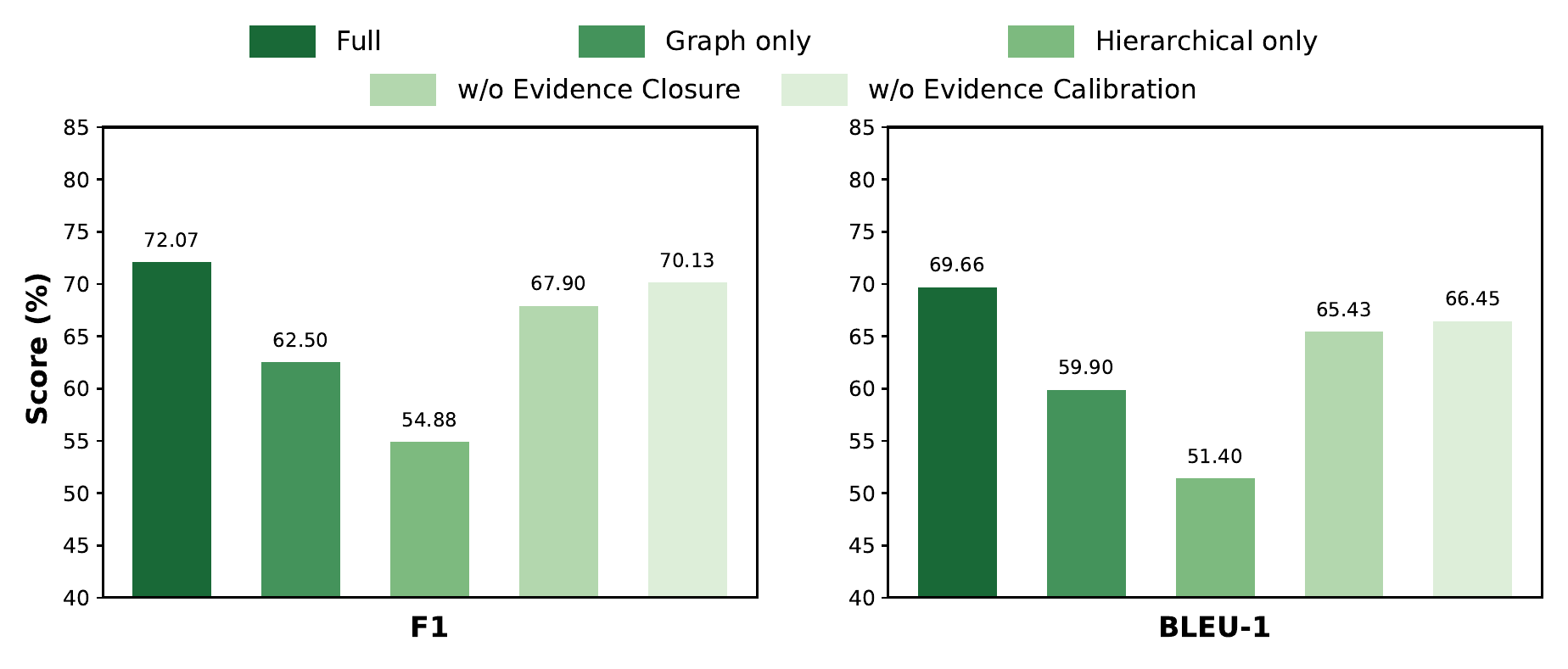}
    \caption{Ablation study on HotpotQA with 56K-token contexts and GPT-4o-mini. The full model outperforms both single-view variants, demonstrating the complementarity of graph-based relational retrieval and hierarchical contextual retrieval. Evidence closure and calibration provide further gains on both F1 and BLEU-1.}
    \label{fig:ablation}
\end{figure}

\subsection{Ablation Study}
Figure ~\ref{fig:ablation} reports ablation results on HotpotQA with 56K-token contexts and GPT-4o-mini. We compare the full model with single-view variants and variants without evidence closure or calibration, while keeping all other settings fixed. The full model achieves 72.07 F1 and 69.66 BLEU-1. Retaining only the graph view reduces the scores to 62.50 and 59.90, whereas retaining only the hierarchical view yields 54.88 and 51.40. The stronger graph-only performance is consistent with HotpotQA's emphasis on relational and cross-document reasoning. However, both variants remain substantially below the full model, showing that the two structures provide complementary evidence: the graph connects information distributed across documents, while the hierarchy preserves local and multi-granular context needed to interpret those connections. Removing evidence closure results in 67.90 F1 and 65.43 BLEU-1, while removing evidence calibration yields 70.13 and 66.45. The consistent declines support their roles in completing and refining the evidence returned by dual-view retrieval. Overall, the results demonstrate the importance of combining graph and hierarchical retrieval, with evidence closure and evidence calibration providing further support to the retrieved evidence.



\begin{figure}[t]
    \centering
    \includegraphics[width=\linewidth]{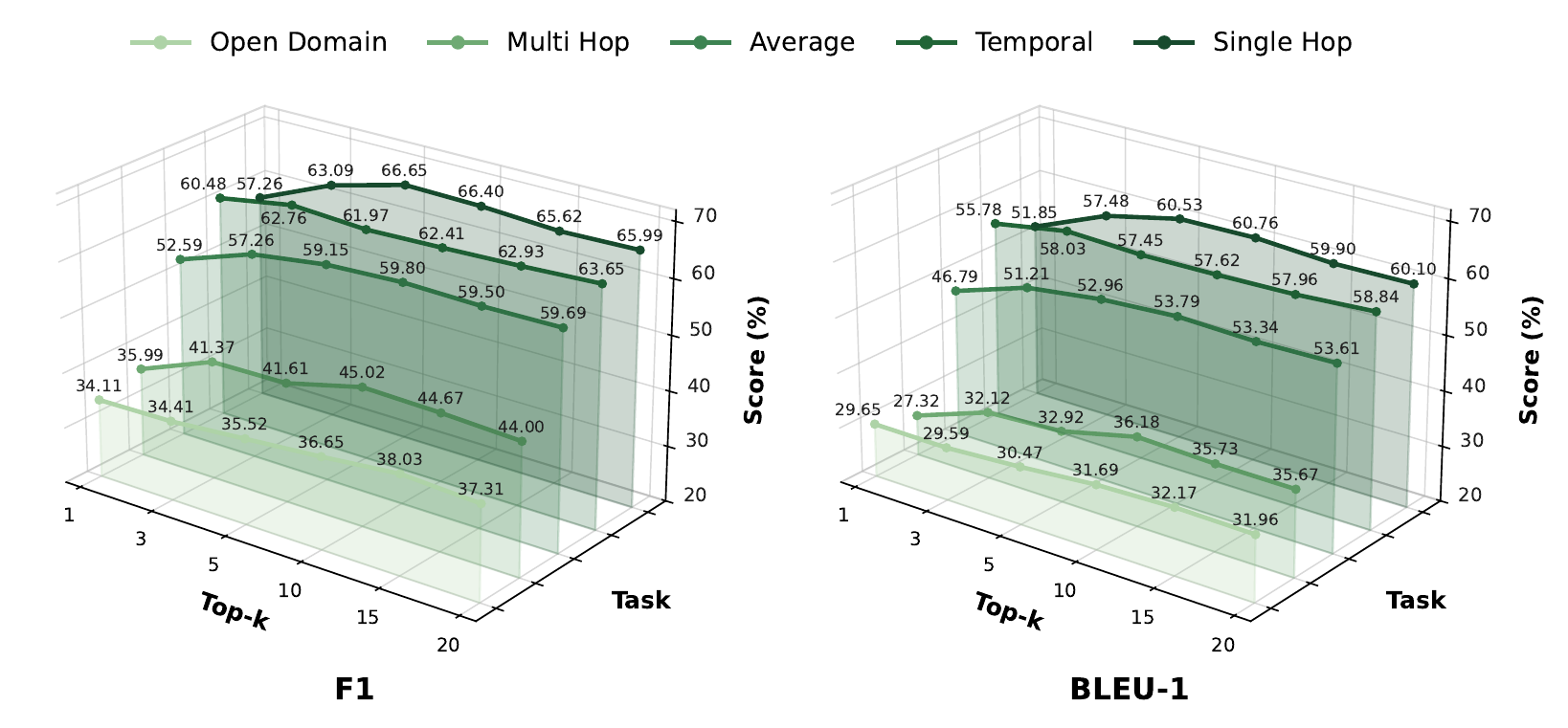}
    \caption{Effect of the retrieval budget on LoCoMo with GPT-4o-mini. Performance improves markedly from top-1 to top-5, reaches its best average at top-10, and remains stable under larger budgets. We use top-5 in the main experiments to match the retrieval setting of all baselines.}
    \label{fig:budget}
\end{figure}

\subsection{Effect of the Retrieval Budget}
Figure~\ref{fig:budget} examines the sensitivity of Zero-Mem to the retrieval budget, defined as the maximum number of primary candidates retained in $M(q)$ before evidence closure. Increasing $Top-K$ from 1 to 5 substantially improves the average F1 and BLEU-1 scores from 52.59 and 46.79 to 59.15 and 52.96, respectively. Performance reaches its highest overall level at $Top-10$, while larger budgets yield only minor fluctuations, indicating diminishing returns from additional evidence. The task-wise results exhibit different saturation points: single-hop questions require relatively few candidates, whereas multi-hop, temporal, and open-domain questions generally benefit from broader evidence coverage. Overall, Zero-Mem remains stable across moderate retrieval budgets. We use $Top-5$ in the main experiments to match the retrieval setting of all baselines; this configuration trails $Top-10$ by only 0.65 F1 and 0.83 BLEU-1 while retaining half as many primary candidates.



\section{Conclusion}
We introduced Zero-Mem and formalized zero-token memory operations, an operating regime in which every operation outside final question answering invokes no LLM and consumes no LLM input or output tokens. Zero-Mem preserves original interaction traces and retrieves evidence through complementary relational and temporally ordered views without generating intermediate memory representations. Comprehensive experiments demonstrate competitive performance across long-term conversational memory and long-context multi-hop reasoning. Ablations further confirm the complementarity of the two evidence views. With an identical final-QA reader and an equivalent context budget, Zero-Mem eliminates memory-operation token consumption and reduces latency by 57.6\% relative to the most time-efficient baseline. These results show that effective agent memory does not require generated intermediate representations and establish provenance-preserving evidence selection as a practical alternative to generative memory pipelines.

\bibliography{aaai2027}


\end{document}